\documentclass[letterpaper, 10 pt, conference]{ieeeconf}  

\IEEEoverridecommandlockouts                              

\usepackage{bm}
\usepackage[pdftex]{graphicx}
\usepackage{graphics} 
\usepackage{epsfig} 
\usepackage{mathptmx} 
\usepackage{times} 
\usepackage{amsmath} 
\usepackage{amssymb}  
\usepackage{algorithm}
\usepackage[noend]{algpseudocode}
\usepackage{multirow}
\usepackage{tikz}
\usepackage{lettrine}
\usepackage{color}
\usepackage{hyperref}

\newboolean{highlight}
\setboolean{highlight}{false}
\newcommand{\change}[1]{\ifthenelse{\boolean{highlight}}{\textcolor{red}{#1}}{#1}}

\newcommand{\figref}[1]{Fig.~\ref{#1}}

\usepackage{cuted}
\usepackage{capt-of}

\title{\bf Dense Tactile Force Estimation using GelSlim and inverse FEM}

\author{
  \authorblockN{Daolin Ma{*}, Elliott Donlon{*}, Siyuan Dong, Alberto Rodriguez} 
  \authorblockA{
     Mechanical Engineering Department --- Massachusetts Institute of Technology\\
    {\tt\small <daolinma,edonlon,sydong,albertor>@mit.edu}} 
\thanks{This work was supported by NSF award [IIS-1637753] through the National Robotics Initiative and by the Amazon Research Awards. This article reflects the opinions and conclusions of its authors and not Amazon.} %
\thanks{* Authors with equal contribution.}
  }

\begin{document}
\maketitle
$ $
\thispagestyle{empty}
\pagestyle{empty}

\begin{strip}
\centering
\vspace{-28mm}
\includegraphics[width=\textwidth]{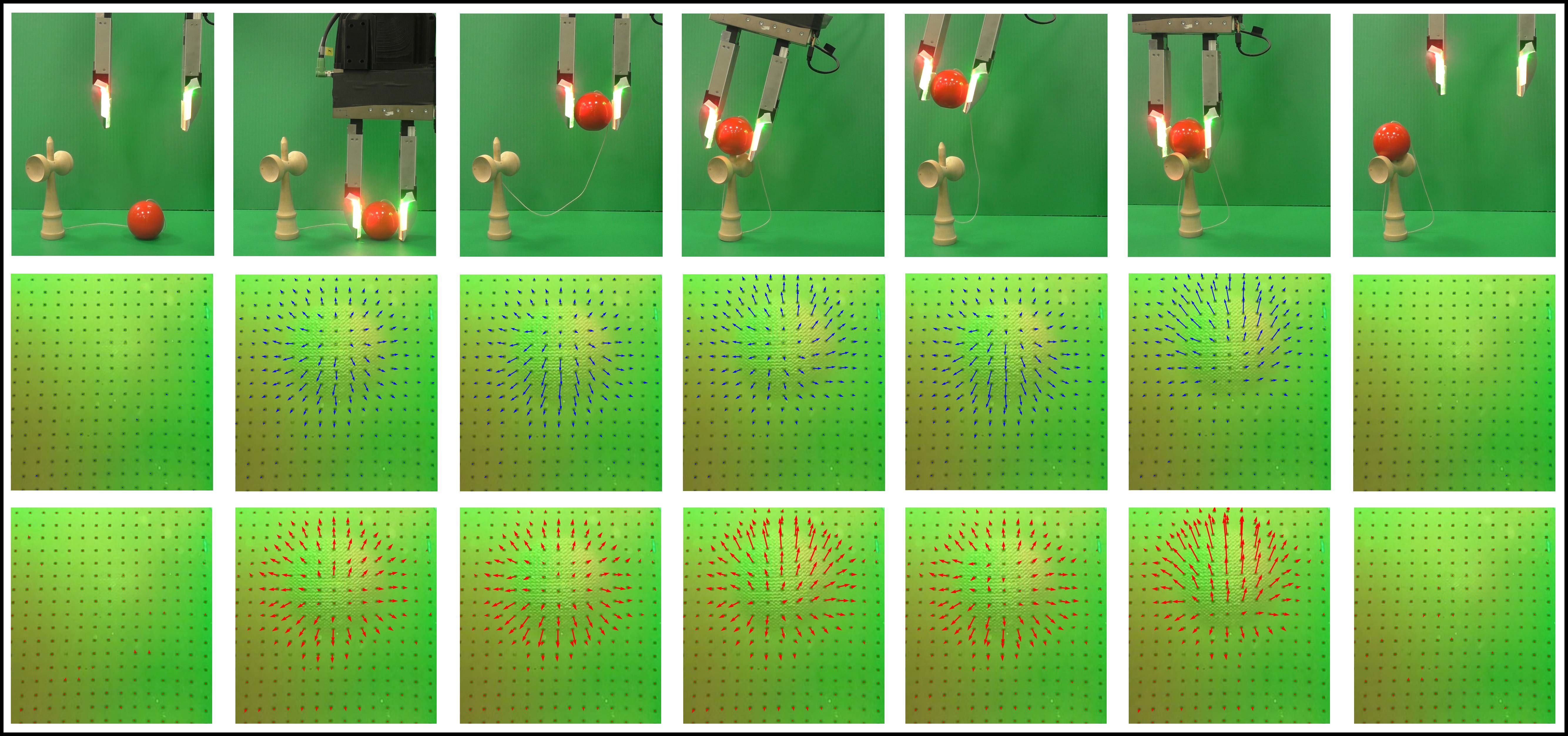}
A sequence of Kendama manipulations with corresponding displacement field (yellow) and force field (red). Video can be found on Youtube: \href{https://youtu.be/hWw9A0ZBZuU}{https://youtu.be/hWw9A0ZBZuU}
\label{fig:feature-graphic}
\end{strip}

\begin{abstract}

In this paper, we present a new version of tactile sensor GelSlim 2.0 with the capability to estimate the contact force distribution in real time. The sensor is vision-based and uses an array of markers to track deformations on a gel pad due to contact. A new hardware design makes the sensor more rugged, parametrically adjustable and improves illumination. Leveraging the sensor's increased functionality, we propose to use inverse Finite Element Method (iFEM), a numerical method to reconstruct the contact force distribution based on marker displacements. The sensor is able to provide force distribution of contact with high spatial density. Experiments and comparison with ground truth show that the reconstructed force distribution is physically reasonable with good accuracy.

\end{abstract}

\IEEEpeerreviewmaketitle

\section{Introduction}

Force feedback is key to humans' amazingly dexterous manipulation skill. Open-loop or vision-guided motion can be inaccurate and slow for many manipulation tasks\cite{Gentilucci1997}. Enabling tactile sensing and tactile-based control in robotic manipulation can significantly improve the robustness, precision and reliability of robot performance. 

Furthermore, many manipulation tasks are often convenient to think about in force space, making contact force an important form of feedback. For example, when picking a ball and placing it on a Kendama, some specific instants are critical phase transitions in the actions. Force information seamlessly captures when successful grasping, lifting, reorienting, placing and releasing of the ball is achieved. Such information is critical for tackling more complex robotic manipulation tasks.



Tactile-sensing technology has been developed in the past to meet this need. Among different technologies for tactile sensing, vision-based tactile sensors, such as GelSight (which uses a camera to track markers on a deformable silicone gel pad) \cite{GelSight_Dong} or GelSlim~\cite{Donlon2018}, are promising due to their high-resolution, convenient data-multiplexing and embedded compliance. 
However, research on vision-based tactile sensors~\cite{Bristol2009,ito2014contact,guo2016measurement,maekawa1993finger,Bauza2018} has focused more intensely on measuring geometry instead of on force reconstruction. 

The key to force reconstruction on vision-based sensors is to accurately model the mapping between force and deformation of the deformable skin. Although force is proportional to deformation of an elastic element, the distribution of forces has a more complex relationship with the distribution of displacements. Indeed, stress and strain are linearly related for a given material according to the theory of elasticity but force and displacement are also geometry-dependent. 

\begin{figure}[t]
    \begin{minipage}[t]{0.48\linewidth}
    \centering
    \includegraphics[width=\linewidth]{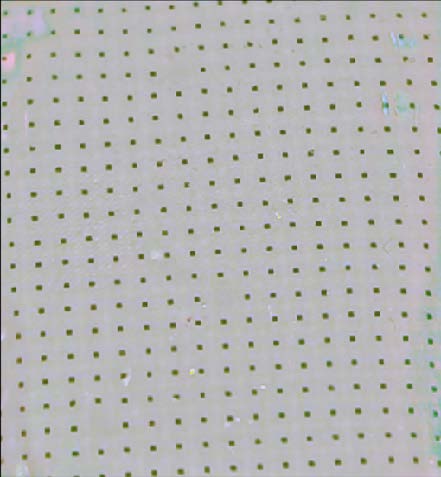}
    (a)
    \end{minipage}%
    \hspace{2mm}
    \begin{minipage}[t]{0.48\linewidth}
    \centering
    \includegraphics[width=\linewidth]{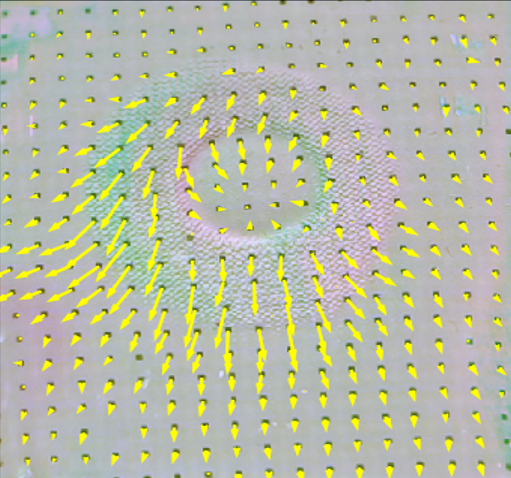}
    (b)
    \end{minipage}%
  \caption{The calibrated image of displacement markers before (a) and after (b) contact on {GelSlim 2.0} sensor with a circular ring pushed in the upper-right direction. Displacements of markers are shown with yellow arrows. Although the contact forces should happen only in the area where contacts are made, the marker tracking algorithm shows that markers in larger area are displaced.}
  \label{fig:displacements of markers}
\end{figure}

When force reconstruction is performed under the erroneous assumption that the force on a point is proportional to its own displacement, we get physically unrealistic predictions. For example, in \figref{fig:displacements of markers}, displacements of markers are observed in areas outside the contact patch. Here, the deformation marker movement outside contact patch is caused by forces internal to the silicone gel pad instead of external loading force. 

The ability to accurately model the force-displacement relationship and reconstruct the force distribution of the entire body simultaneously can greatly improve the efficacy of vision-based tactile sensors. In this work, we propose a numerical approach: inverse Finite Element Method (iFEM), to estimate the force distribution with deformation measured by vision. We also report on the development of GelSlim 2.0 -- a new version of the vision-based tactile sensor with force distribution estimation. Experiments and validation are carried out with GelSlim 2.0.


\section{Related work}

In this section, we review different technologies that allow measurement of force distributions.
\paragraph{Vision-based Tactile Sensors}
Vision-based tactile sensors have the advantage of high spatial resolution and are commonly used to sense contact location and/or texture and/or geometry of the contact object~\cite{GelSight_greensensor,Ferrier2000}. Since force information cannot be directly observed from the contact imprint, researchers have explored different methods to infer force information. Begej built a vision-based tactile sensor that utilized frustrated total internal reflection phenomenon to produce a grey-scale tactile image \cite{begej1988planar}. The sensor can approximately estimate the normal force by calibrating a look-up table between pixel intensity and normal force. Ohka~\textit{et al}. built an optical three-axis tactile sensor that could measure the total forces in 3 axis by observing the variance of contacting area between the conical feeler on the sensor surface and the inner acrylic board \cite{ohka2004sensing}. One popular method of force measurement with vision-based tactile sensors is to track the displacement of a field of markers printed on the elastic sensing surface. The GelSight sensor has black markers distributed on the sensor surface, which move along with the applied external forces \cite{GelSight_Dong}. Yuan~\textit{et al}. use a deep learning method to learn the total three-axis forces and z-axis torque directly from this output~\cite{GelSight_review}. However, the measured forces and torques are noisy and are affected largely by the contact textures. 
To the authors' knowledge, the GelForce sensor is the only vision-based tactile sensor that can measure the distribution of three-axis forces \cite{Gelforce1,Gelforce2}. It uses two layers of markers with different colors to track the motion of the elastic gel surface. With the displacements of the markers, the three-axis forces at each marker position are calculated analytically according to elastostatic theory. The drawback of this method is that it relies on the strong assumption that the elastic surface is semi-infinite. This results in the core equation that a force vector acting on a single marker is linearly dependent only on the displacement vector of that particular marker, which is not valid in general cases. In this paper, we employ the iFEM method, also based on elastostatic theory to establish a global relationship between the field of marker motions and the field of applied forces. The algorithm works for a diverse set of elastic membranes and configurations and works in real time. 

\paragraph{Force Transduction Arrays}
There are many force sensor arrays with various working principles developed to measure force distributions~\cite{engel2003development,papakostas2002large,kim2012transparent}. However, most of them can only sense the distribution of normal force\cite{mukai2004soft,lazzarini1995tactile,lazzarini1995tactile} with spatial resolution of about $2mm$. 
An 8 $\times$ 8 flexible capacitive tactile sensor array built by Lee~\textit{et al}. could measure both normal and shear force distribution \cite{lee2008normal}. Four capacitors in the sensor formed a cell to decompose the contact force into normal and shear components. The sensor also worked with 2 mm spatial resolution. Compared to the potential of optical-based sensors, the sensors discussed above have lower spatial resolution and much smaller amount of sensing elements, but often feature higher bandwidth.



\section{Hardware: GelSlim v2}
In order to enable force distribution estimation, we augment the hardware design of GelSlim~\cite{Donlon2018}.

\subsection{Markers to track displacement}
\begin{figure}[t]
\centering
  \includegraphics[width=0.85\linewidth]{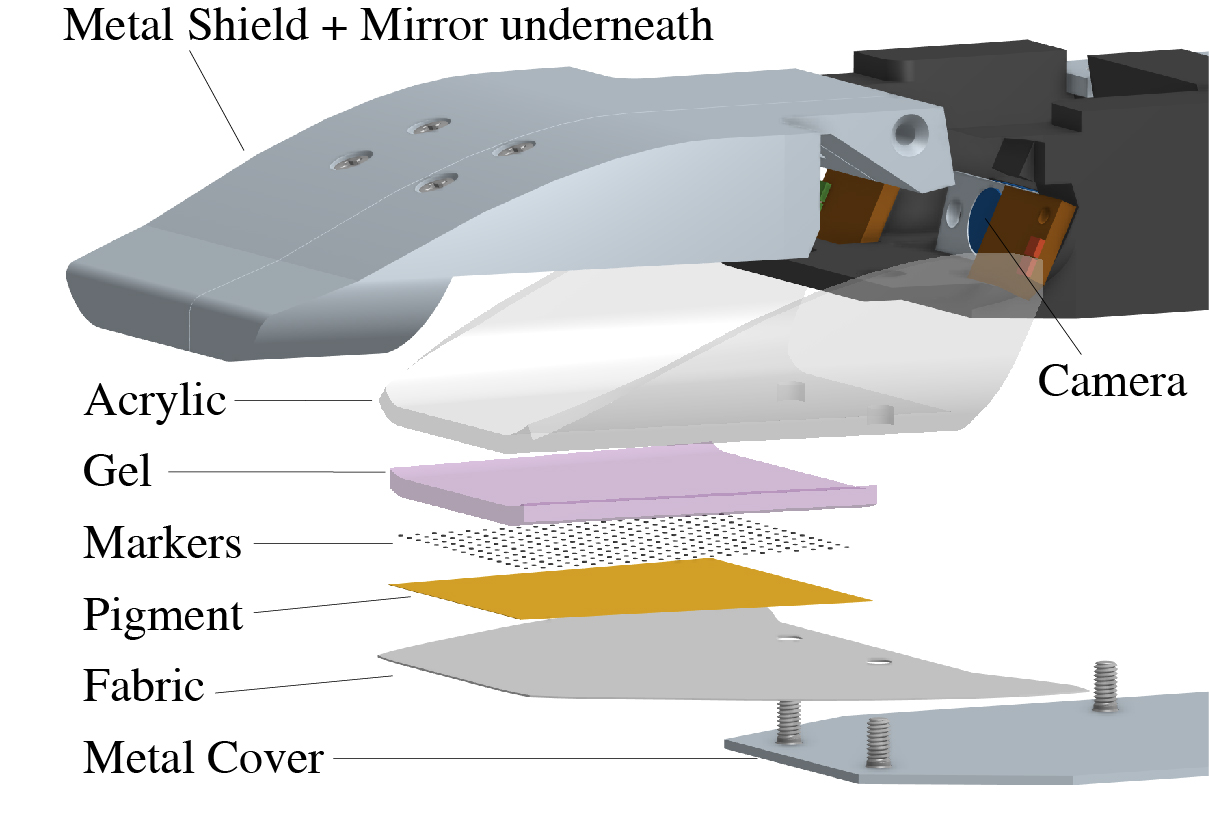}
  \caption{Exploded view of finger construction and maker placement.}
  \label{fig:Explodedview}
\end{figure}

The previous version of the GelSlim sensor was able to measure contact at fairly high resolution using a Raspberry Pi Spy Camera. However, it was difficult to disambiguate shear of the gel pad versus an object sliding across the gel by observing motion of the contact. We solve this problem, and enable real-time iFEM, by adding displacement markers to the gel surface, shown in figure \figref{fig:Explodedview}. These markers are very similar to the ones used in the GelSight sensor \cite{gelsightShear} except that they are a regular grid, which is directly-correlated with nodes of the FEM, instead of a pseudorandom pattern. The markers are printed directly on the outer surface of the gel, under the paint, which enables direct measurement of gel displacement in camera frame. We then use the measured displacement field and knowledge the gel's material properties to make a dense estimation of the contact force field.

\subsection{Improved illumination}
GelSlim's signal strength depends on the visual contrast between contact and non-contact scenarios. For example, when the gel is illuminated with grazing light, a contact region changes the surface normal to reflect more or less light back to the camera. Because of this, the illuminated side appears extra bright and the opposite side appears extra dark. In GelSlim 2.0, we increase signal contrast with dual-color illumination relative to the white illumination used in the previous version. In this case, the gel pad is illuminated with red grazing light from the left and green grazing light from the right. These colors are easy to separate by isolating two of the camera's three (RGB) channels. Therefore, contact information is encoded by differences between channels and the different shadows they cast on the contact surface. These particular LEDs (\textit{LUXEON CZ}) are chosen because their spectral emissions were well-matched with the detection bands of the Raspberry Pi Spy Camera's CCD while also being far apart enough spectrally to be non-overlapping. 

Signal strength also relies on having enough illumination. Vision-based tactile sensors can only run at maximum frame-rate if sufficiently illuminated. To increase the efficiency of the optical path from light source to gel, we rely on Total Internal Reflection (TIR) through a curved light guide to reduce the number of mirror reflections. Relative to the previous design that used a hard 45$^{\circ}$ mirror reflection to route light to the gel, GelSlim 2.0 features a curve that maintains TIR to increase brightness and therefore frame-rate. 

\subsection{More rugged}
One practical requirement for a robotic tactile sensor is its durability. The previous design was able to sustain the wear of thousands of grasps with relatively little functional degradation. Much of this degradation experienced was due to damage of the optical path. In GelSlim 2.0 we have redesigned the finger to have not only stronger components, but also pre-compression in the optical path to prevent tensile failure. This allows the sensor to continue its useful life with less wear.


\subsection{Parametric design}
\begin{figure}[t]
\centering
  \includegraphics[width=0.85\linewidth]{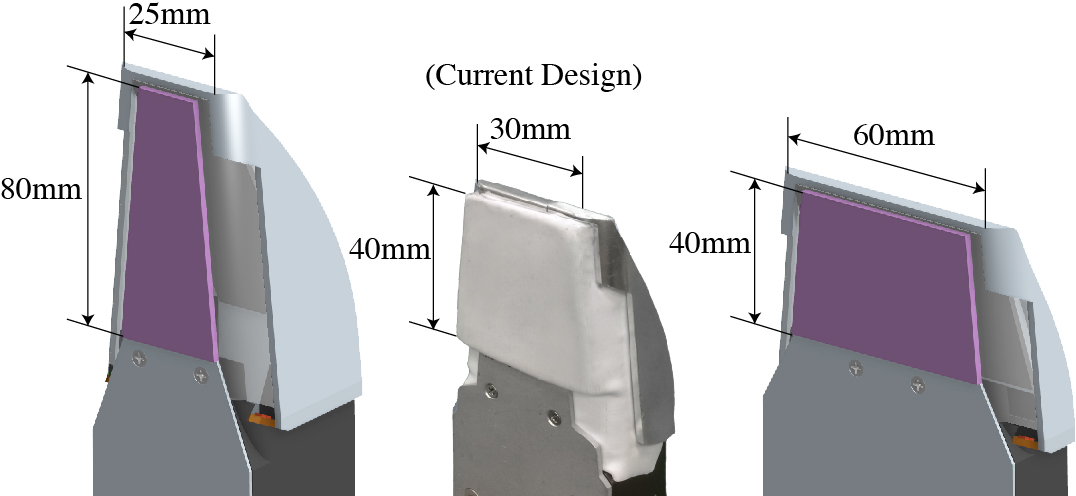}
  \caption{A family of GelSlim sensors generated from the same parametric design.}
  \label{fig:Parametric design}
\end{figure}

Our design is parameterized using three sets of variables: 
\begin{enumerate}
    \item \textbf{Camera parameters} like depth of field and maximum viewing angle are assumed to be given based on specific camera hardware.
    \item \textbf{Independent design variables} like gel size are set by the designer based on the requirements of the task. 
    \item \textbf{Dependent design variables} like finger thickness and overall length are driven by each of the independent variable sets listed above.
\end{enumerate}
GelSlim 2.0 uses CAD that is fully-parameterized in \textit{Onshape} by these variables (\figref{fig:Parametric design}). This parametric design makes it easy to scale for various manipulation tasks and optical systems. Sensors of this current version can be made with gel pads down to 20 mm x 20 mm and finger thickness down to 18 mm. Further decreasing thickness intensifies the keystone warp of the image and requires a large depth of field. Cameras with larger depth of field or folded optical paths could could enable thinner GelSlim sensors.

\section{New capability: Force reconstruction}
The key to the relationship between displacement and force fields is an FEM model of the gel. Therefore, FEM is first briefly reviewed in this section. Then we'll show how to calculate the force distribution based on the displacement of markers. Finally, we describe the pipeline for performing force reconstruction with iFEM on vision-based tactile sensor.

\subsection{FEM and Hex-8 element}

\begin{figure}[hbtp]
\centering
\vspace{-4pt}
  \includegraphics[width=0.8\linewidth]{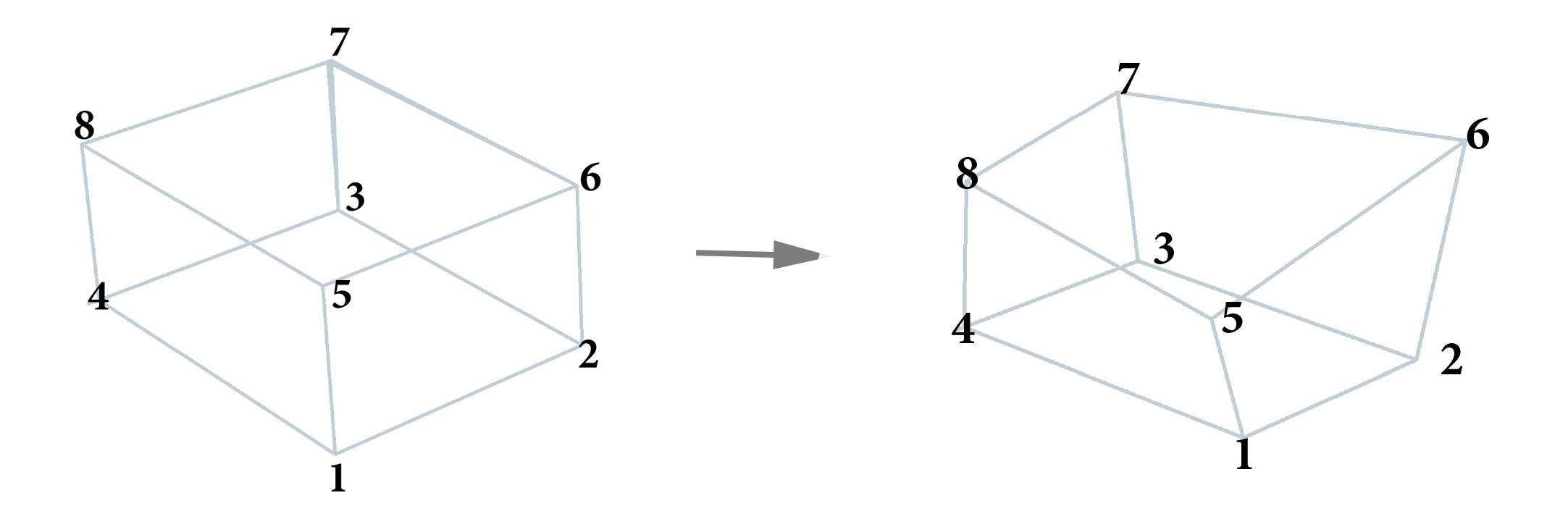}
  \caption{Sketch of a Hex-8 element with its 8 nodes before (left) and after (right) deformation.}
  \label{fig:hex-8}
\end{figure}

The essence of FEM is to discretize an object into small and simple elements, whose force and deformation relationship are well defined by the theory of elasticity. 
In this paper, we employ Hex-8 elements to discretize the gel pad, for reason of both accuracy and efficiency. As shown in Fig.~\ref{fig:hex-8}, one Hex-8 element includes 8 nodes and has 24 degrees of freedom (DoFs). The combination of displacements of all 8 nodes describes the deformation of one FEM element. The external force at each node is linearly dependent on all displacements at all 8 nodes within the element though an element stiffness matrix. The elements are then constrained to each other through boundary conditions and assembled together with a single high-dimensional stiffness matrix. 

\subsection{Force vs. deformation: Stiffness Matrix}

\begin{figure}[t]
\centering
  \includegraphics[width=0.8\linewidth]{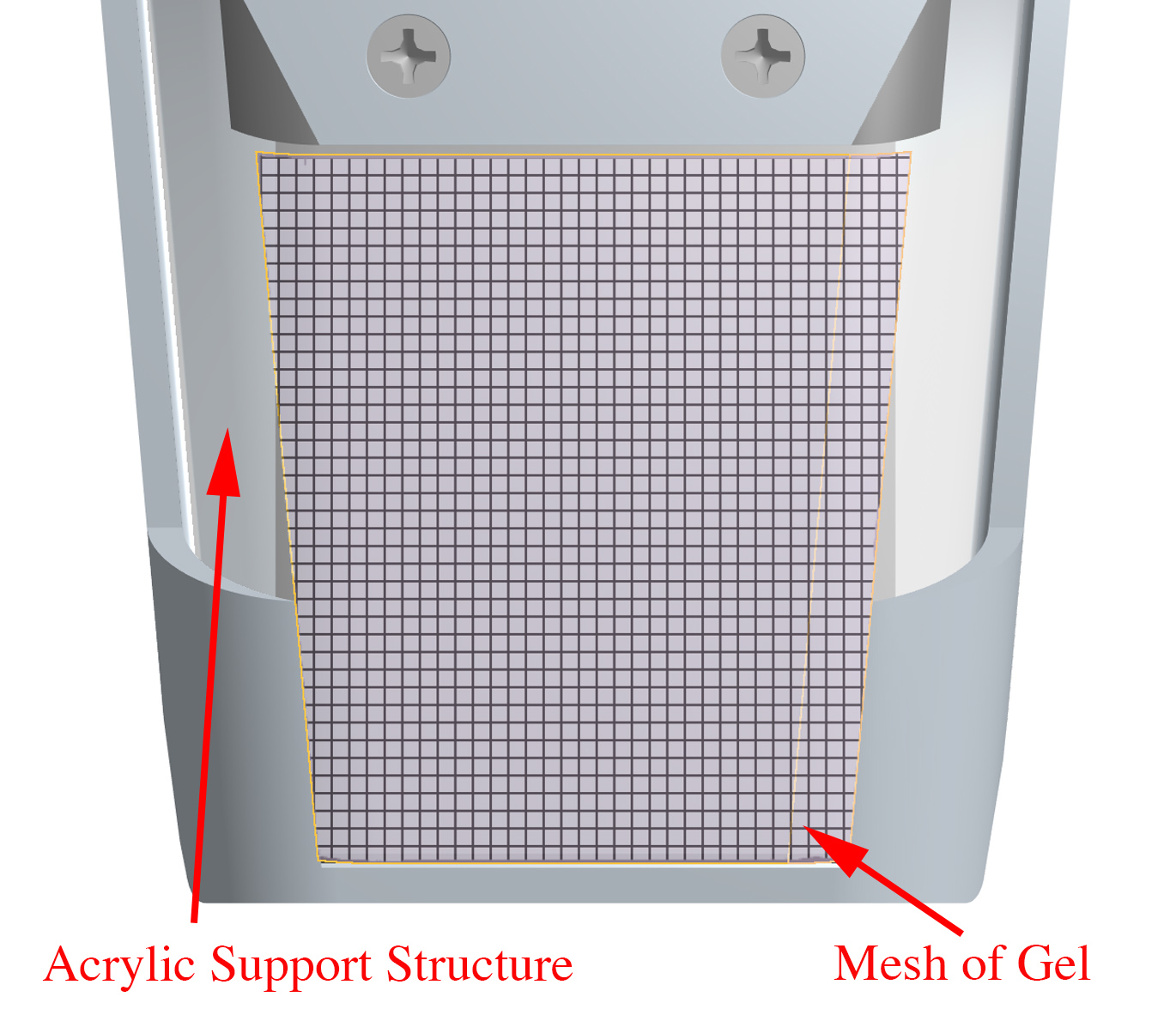}
  \caption{The silicone gel pad (purple) is meshed into one layer of Hex-8 elements. The full mesh includes 1964 elements with 3698 nodes. }
  \label{fig:meshed gel pad}
\end{figure}

The first step to force estimation is to discretize the object's 3D geometry into $m$ 8-node hexahedron elements, as in Fig.\ref{fig:meshed gel pad}, with a total of $n$ nodes. The mesh should cover the entire visible area of the gel. One could mesh the shape of gel exactly with elements that are non-uniform in shape and size. But for easy implementation, we mesh on an area a little larger than the bounding box of the gel with uniform sized elements and then crop it to the gel's actual size. 

With FEM, all external forces are assumed to be applied at the nodes. Similarly, we will solve the contact forces on the nodes. If we denote the displacement of all nodes as a displacement vector $\bm{U} = (\delta_x^1,\delta_y^1,\delta_z^1,\delta_x^2, \delta_y^2,\delta_z^2,...,\delta_{z}^n)$, then the external force vector $\bm{F} = (f_x^1, f_y^1,f_z^1,f_x^2,f_y^2,f_z^2,...,f_{z}^n)$, which represents externally applied forces at all nodes, would be linearly dependent on the displacement vector $\bm{U}$ as in Equation.~\ref{eqn:Force_dis_law}

\vspace{-6pt}
\begin{equation}
    \bm{F} = \bm{K}\bm{U},
    \label{eqn:Force_dis_law}
    \vspace{-6pt}
\end{equation}

where the $3n \times 3n$ matrix $\bm{K}$ is referred to as the stiffness matrix and can be determined via standard FEM theory with 8-node hexahedron elements \cite{bathe2006finite}. Beside the 3D mesh of the gel pad, the only parameters required to obtain the stiffness matrix are Young's Modulus and Poisson's ratio.Young's Modulus is a parameter that describes the stiffness of the gel, while Poisson's ratio is a measure of the Poisson effect, the phenomenon by which a material tends to expand in directions perpendicular to the direction of compression. Our tensile test at different speeds results suggest that the viscosity effect can be neglected. The two parameters are $147$MPa and $0.3223$ in this case, but they can be sensitive to gel preparation process. 

\subsection{Displacement measurement}
\label{part: marker tracking}

We then implement an image processing algorithm to track the tangential displacements of markers, as shown in Fig.\ref{fig:displacements of markers} (b). The algorithm includes two steps. 1) Find marker locations in the former and current frame. 
2) Match the marker position between the two frames and compile the displacement field. Because of the deformation of the gel, some of markers can be missed by the marker detector so we cannot compute the marker displacements by simply taking the difference of the two position matrices. Instead, we calculate the distances between the target marker in the former frame and the marker with the smallest distance is the potential correspondence and then further check its validity. 

Since the distribution of the detected markers can be not uniform, we interpolate the displacement vector of nodes based on the computed marker displacement field. 


We use a single layer of nodes with fixed boundary condition (gel fixed to the acrylic) to characterize the system, so that the displacement vectors of all nodes are directly observable. Thus we avoid computation efforts to solve un-observable internal nodes. The $z$ direction displacements of markers can be extracted from the deformation of gel in $z$ direction. Although the method can work on arbitrary shapes with precise gel deformation information in the depth direction ($z$ direction), for easy implementation, we restrict our current object set in experiment to objects with simple geometry whose $z$-direction deformation can be constructed by looking at the shape of contact patch like a sphere or a cylinder. 

\subsection{Compensation to projection error}

\begin{figure}[hbtp]
\centering
  \includegraphics[width=0.7\linewidth]{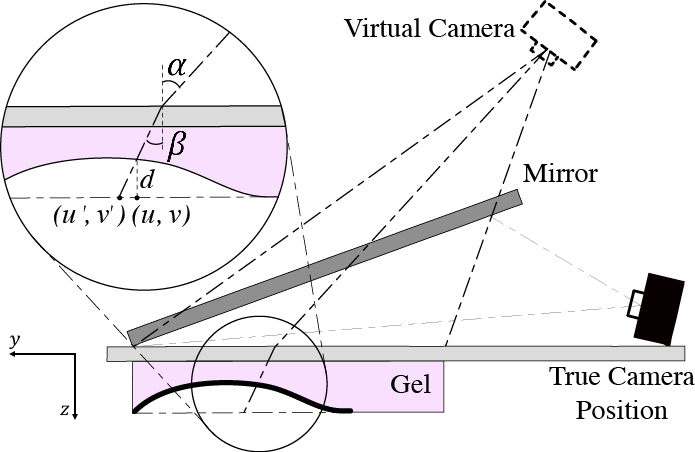}

    
  \caption{A projection error is the displacement error between $(u,v)$ and $(u',v')$. The virtual camera is a projection of the real camera position across the mirror. The optical path at the contact point is expanded for clarity. }
  \label{fig:DepthCompensation}
\end{figure}

Since the camera is not looking at the gel pad perpendicularly, as shown in Fig.~\ref{fig:DepthCompensation}, even though a marker is displaced only in the normal direction, the camera can still observe displacements of markers due to perspective. The displacements of markers observed by camera are caused by both tangential and normal deformation of the gel pad. Therefore, we need to decouple the tangential and normal components in order to retrieve the true 3D displacements of each marker. 

As illustrated in Fig.~\ref{fig:DepthCompensation}, we call projection error to the displacement component seen from camera caused by normal deformation, and denote it as $\delta=(u'-u,v'-v')$. By analyzing the optic path from a marker to virtual camera, we can model the projection error caused by the displacement of the marker in $z$ direction deformation. If we denote $C_v(x_c,y_c,z_c)$ the coordinates of virtual camera in world reference frame, and $M(x,y,z)$ the coordinates of a marker on the gel in world reference frame,  the 2D vector from virtual camera to the marker in gel pad plane is $\bm{r} = (x-x_c,y-y_c)$. Thus the angle of incidence $\alpha$ is determined by:
\begin{equation}
    \alpha = \frac{\left|{\bm{r}}\right|}{\left|{z-z_c}\right|}.
\end{equation}
Then, angle of refraction $\beta$ is 
\begin{equation}
    \beta = \arcsin(\sin(\alpha)/\gamma),
\end{equation}
where $\gamma$ is the refractive index of the silicone gel.

Supposing the $z$ direction marker displacement $d$ is known, the projection error $\Delta$ is
\begin{equation}
    \Delta = \frac{\bm{r}}{\left|\bm{r}\right|} d \tan{\beta}
\end{equation}
The optical refraction in acrylic is ignored when developing these equations because the refractive index of acrylic is close to that of the silicone gel by design. 

\subsection{The algorithm of force estimation}

\begin{figure}[hbtp]
\centering
  \includegraphics[width=\linewidth]{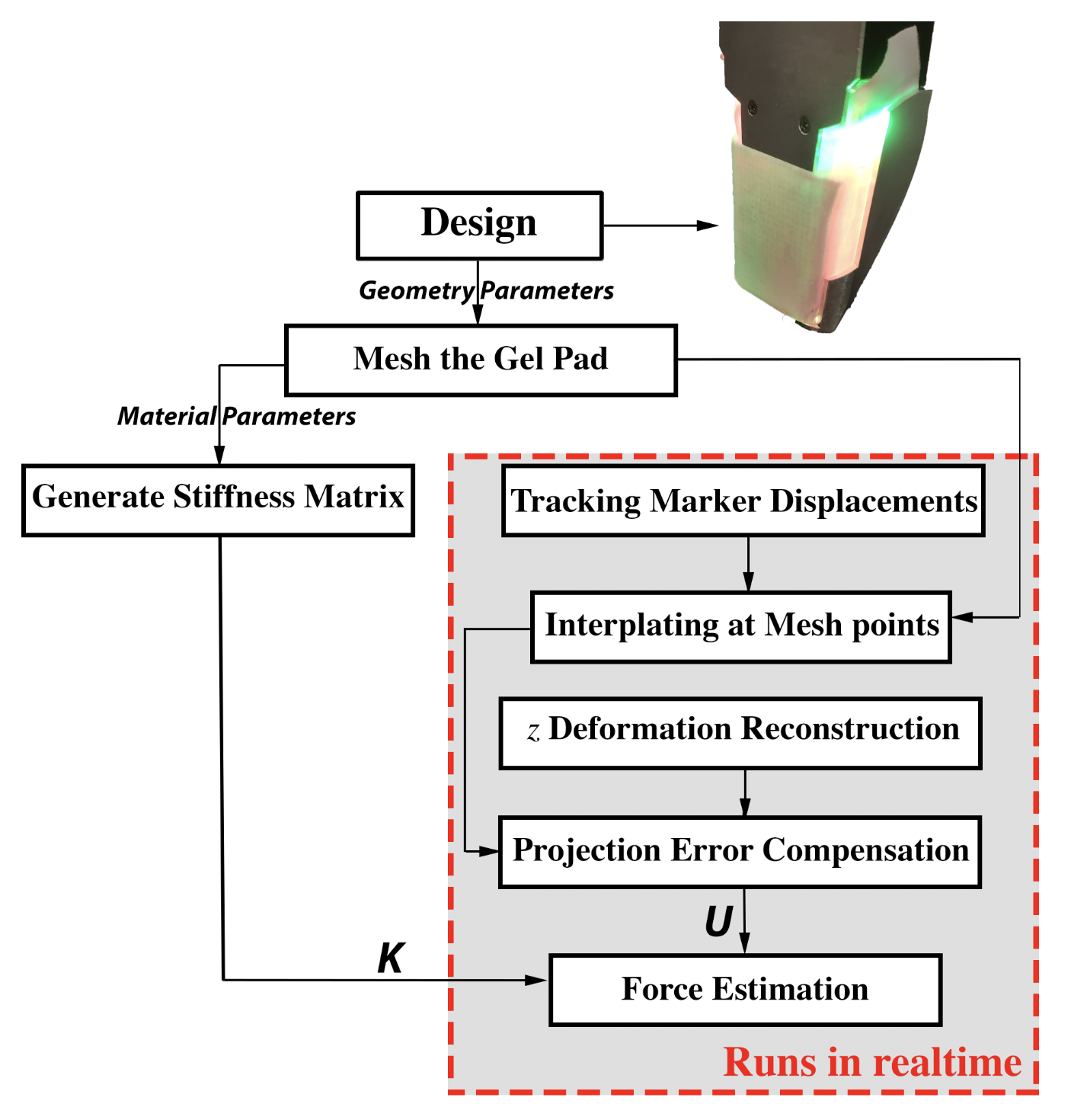}
  \caption{ Force estimation pipeline.}
  \label{fig:force estimation pipeline}
\end{figure}

\figref{fig:force estimation pipeline} summarize the algorithm of force estimation. Generating the stiffness matrix $\bm{K}$ is usually computationally expensive. However, since it is constant for a particular choice of elastic skin configuration, it can be pre-computed and loaded to system off-line. Thus, the force estimation runs in real time.

\section{Experiments and Validation}

We perform an experiment where we push the sensor against a known geometry using an ATI Gamma force-torque sensor to validate force distribution reconstructed with iFEM. 

\begin{figure}[tb]
    \centering
    \includegraphics[width=0.65\linewidth]{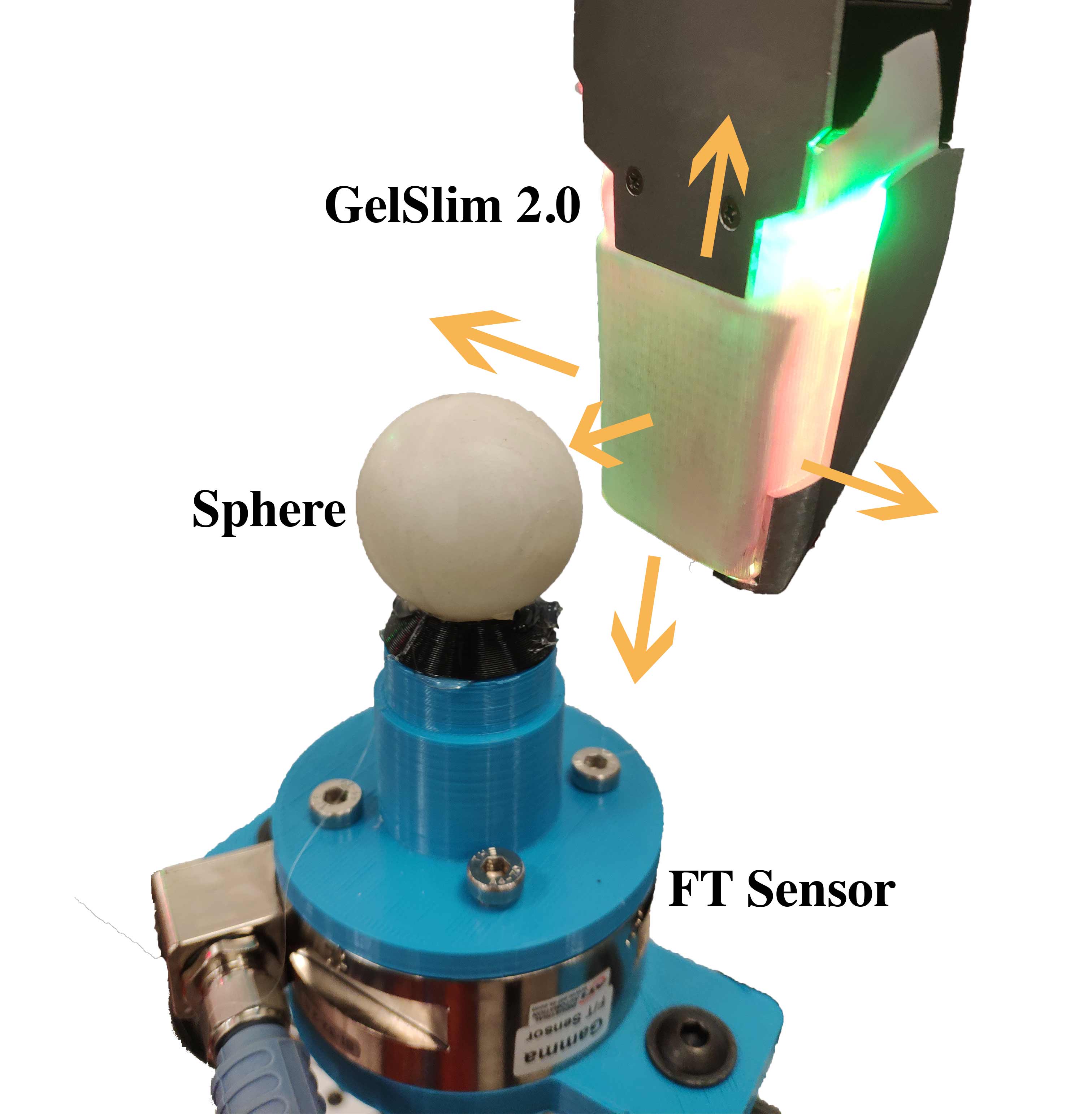}
      \caption{Validation experiment setup.}
    \label{fig:experiment setup}
\end{figure}

\paragraph{Experimental setup}
As shown in ~\figref{fig:experiment setup}, a sphere is installed on a Force/Torque (FT) sensor (ATI Gamma) while the GelSlim 2.0 sensor on a robot hand is programmed to push on the sphere and then slide to different directions. The contact would result in tangential as well as normal forces between GelSlim and the sphere. The FT sensor measures the contact forces and provides ground truth measurement to validate the resultant force measured by the GelSlim sensor calculated from the reconstructed force distribution. 

\paragraph{Tangential force distribution}

\begin{figure}[htbp]
    \begin{minipage}[t]{0.47\linewidth}
    \centering
    \includegraphics[width=\linewidth]{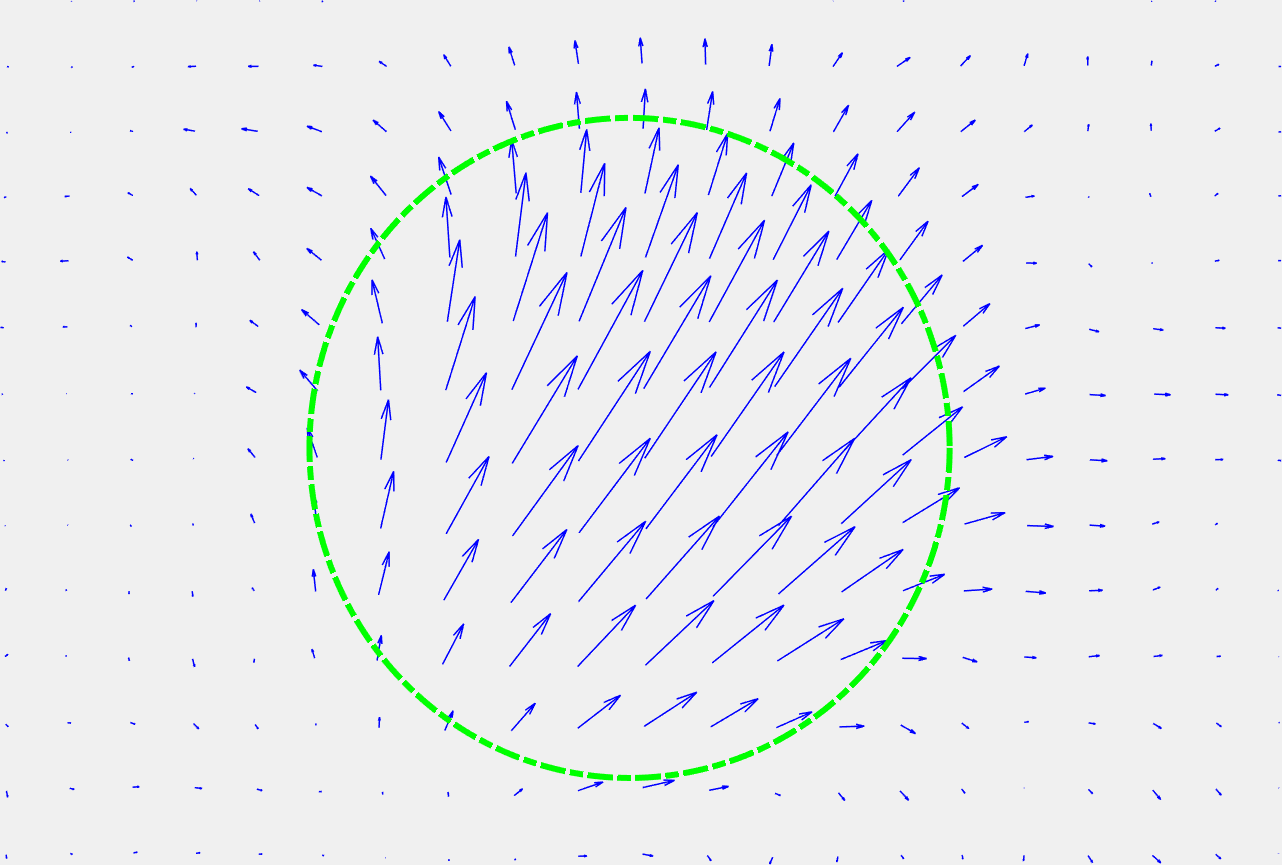}
    \end{minipage}%
    \hspace{2mm}
    \begin{minipage}[t]{0.47\linewidth}
    \centering
    \includegraphics[width=\linewidth]{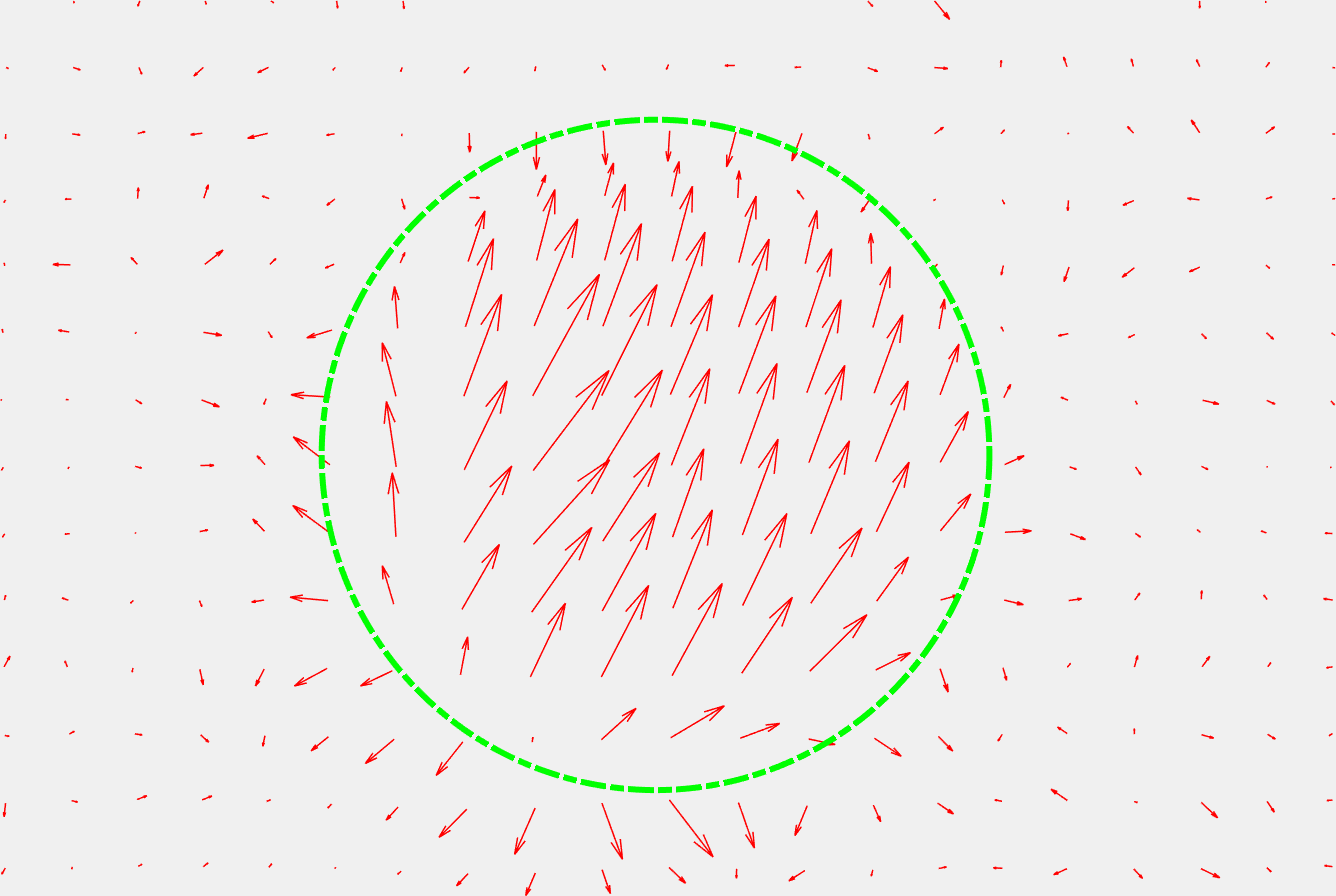}
    \end{minipage}%
  \caption{Comparison between marker displacement field (left, blue) and force distribution field (right, red). The green circles represent the area of contact. It shows that the reconstructed tangential forces are naturally limited in contact patch in most area, despite globally existing noises. The measured force in down-left area is probably caused by the stretched fabric that is pressed into the gel to some degree, which is not caught by our simplified normal displacement reconstruction.}
  \label{fig:Displacementmap_vs_forcemap}
\end{figure}

\figref{fig:Displacementmap_vs_forcemap} shows the displacement field on gel and tangential force field reconstructed with iFEM for an example touch against the sphere in \figref{fig:experiment setup}. The differences between displacement field and force field are two-fold: First, the reconstructed forces are well located inside the contact patch, while displacement vectors extend outside the contact patch. This matches the physical intuition that the gel can only experience reaction forces at contact points while deformations extend though the material. Therefore, the force distribution from iFEM is more physically realistic than directly multiplying displacement distribution by a scalar stiffness. Secondly, the direction of the tangential force at a mesh point is not necessarily the same as its tangential displacement. At certain areas such as boundary of contact, the two directions can be even opposite to each other. Since normal deformation is needed to compute force distribution, for this experiment, we obtain it via analysis of the circular contact patch. We first locate the patch via an image processing algorithm and then calculate the 3D geometry for the known sphere radius.

\paragraph{Normal force distribution}

\begin{figure}[t]
    \centering
    \includegraphics[width=0.8\linewidth]{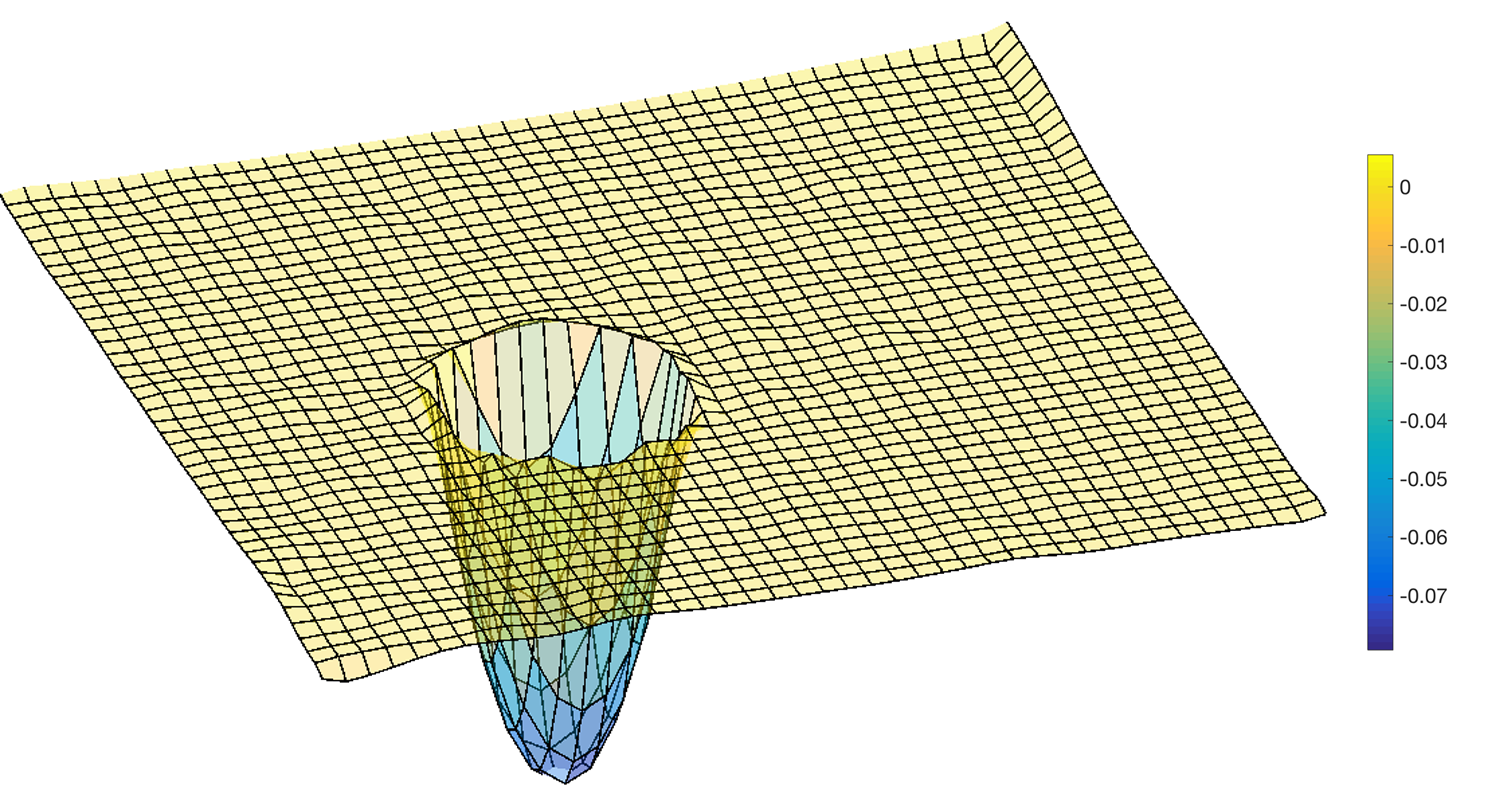}
      \caption{Normal force distribution (unit: N).}
    \label{fig:normal force distribution}
\end{figure}

Although tangential force distribution is what we usually care about, the normal force distribution is also important in some cases. We show the reconstructed normal force distribution of spherical contact in Fig.\ref{fig:normal force distribution}. This figure shows a smoothly-distributed normal force on the surface which is reasonable and consistent with the geometry of a sphere. 

\paragraph{Validation of resultant force }

\begin{figure}[t]
    \begin{minipage}[t]{0.48\linewidth}
    \centering
    (a)
    \includegraphics[width=\linewidth]{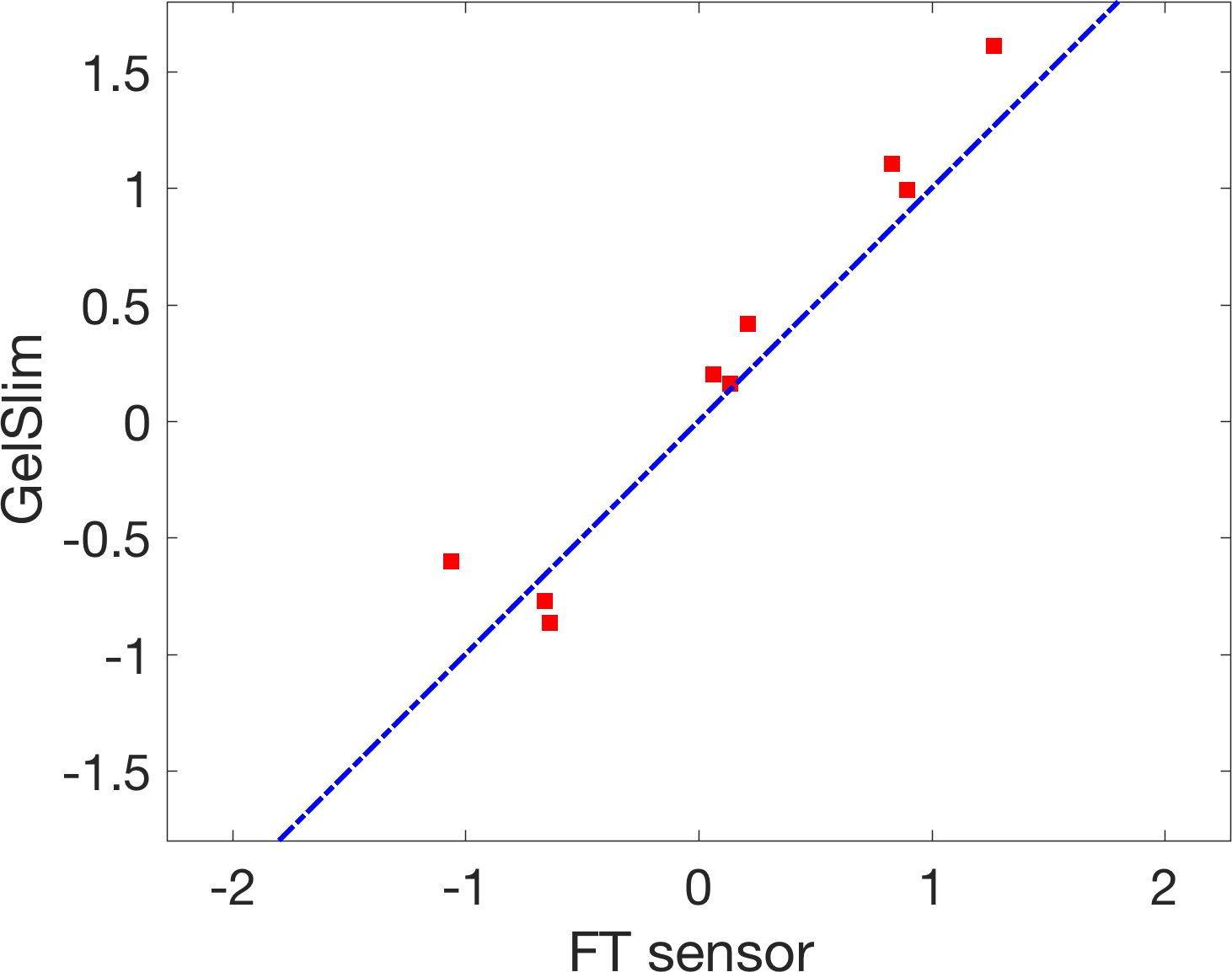}
    \end{minipage}%
    \begin{minipage}[t]{0.48\linewidth}
    \centering
    (b)
    \includegraphics[width=\linewidth]{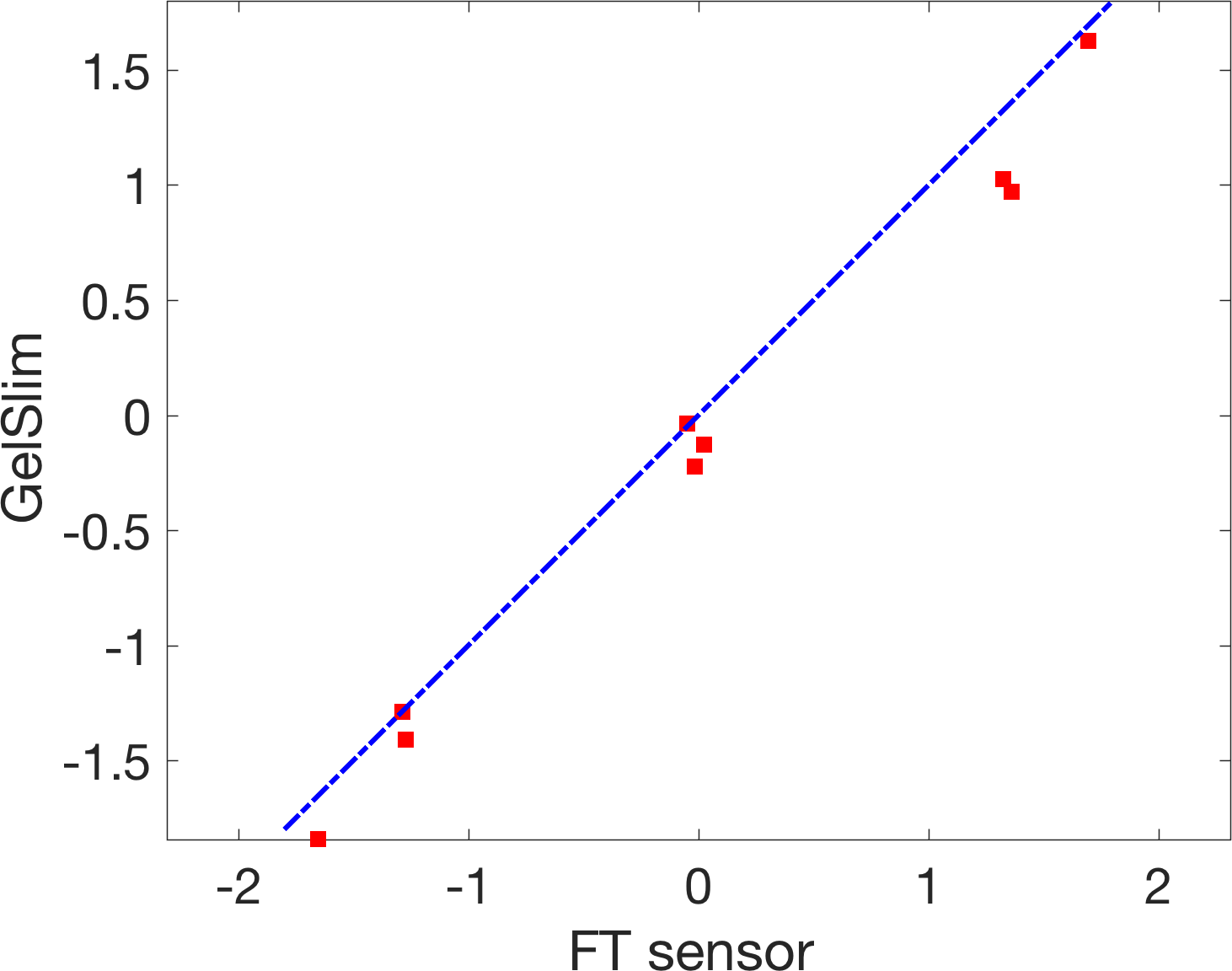}
    \end{minipage}%
    \vspace{10mm}
    \begin{minipage}[t]{0.48\linewidth}
    \centering
    (c)
    \includegraphics[width=\linewidth]{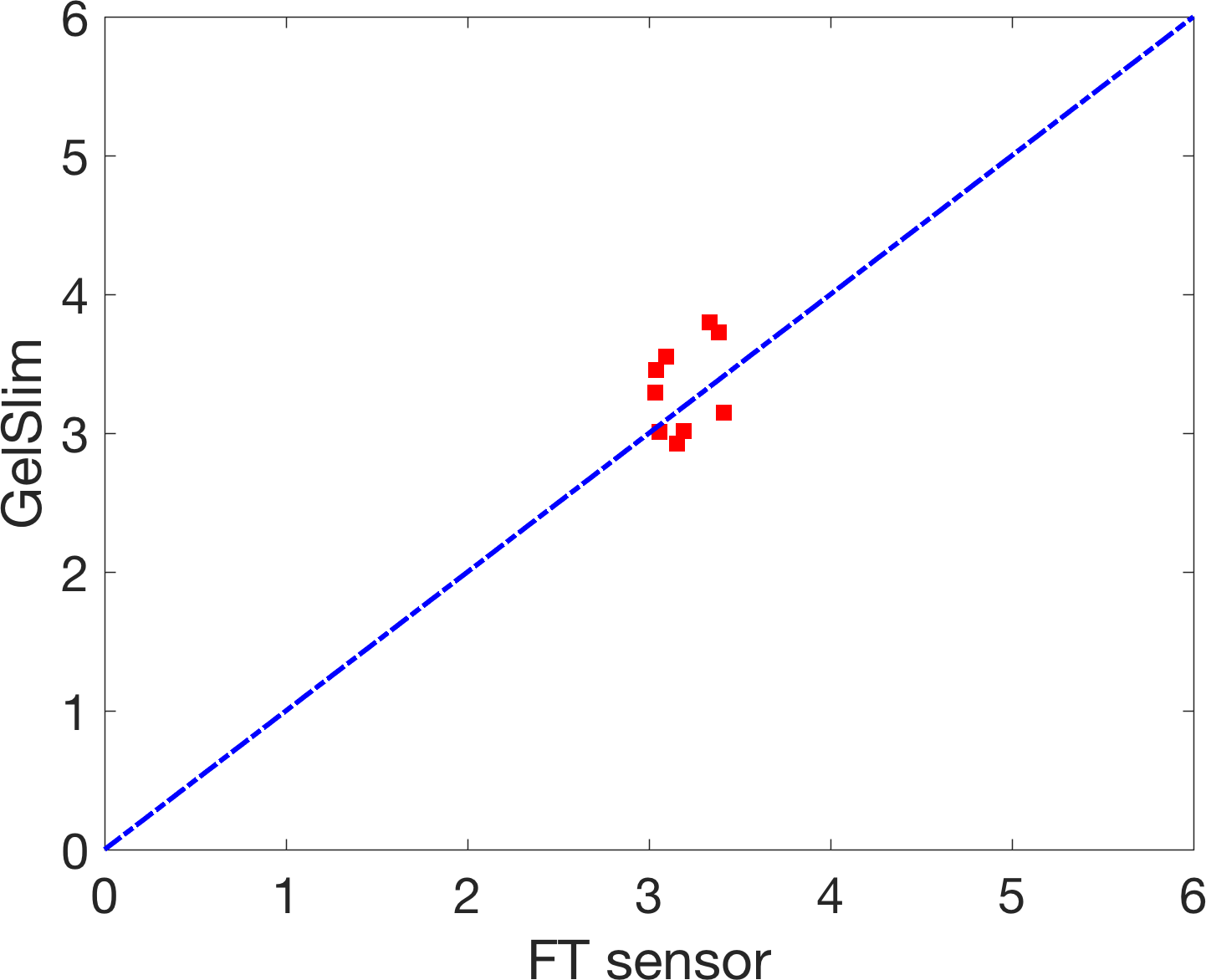}
    \end{minipage}%
    \begin{minipage}[t]{0.48\linewidth}
    \centering
    (d)
    \includegraphics[width=\linewidth]{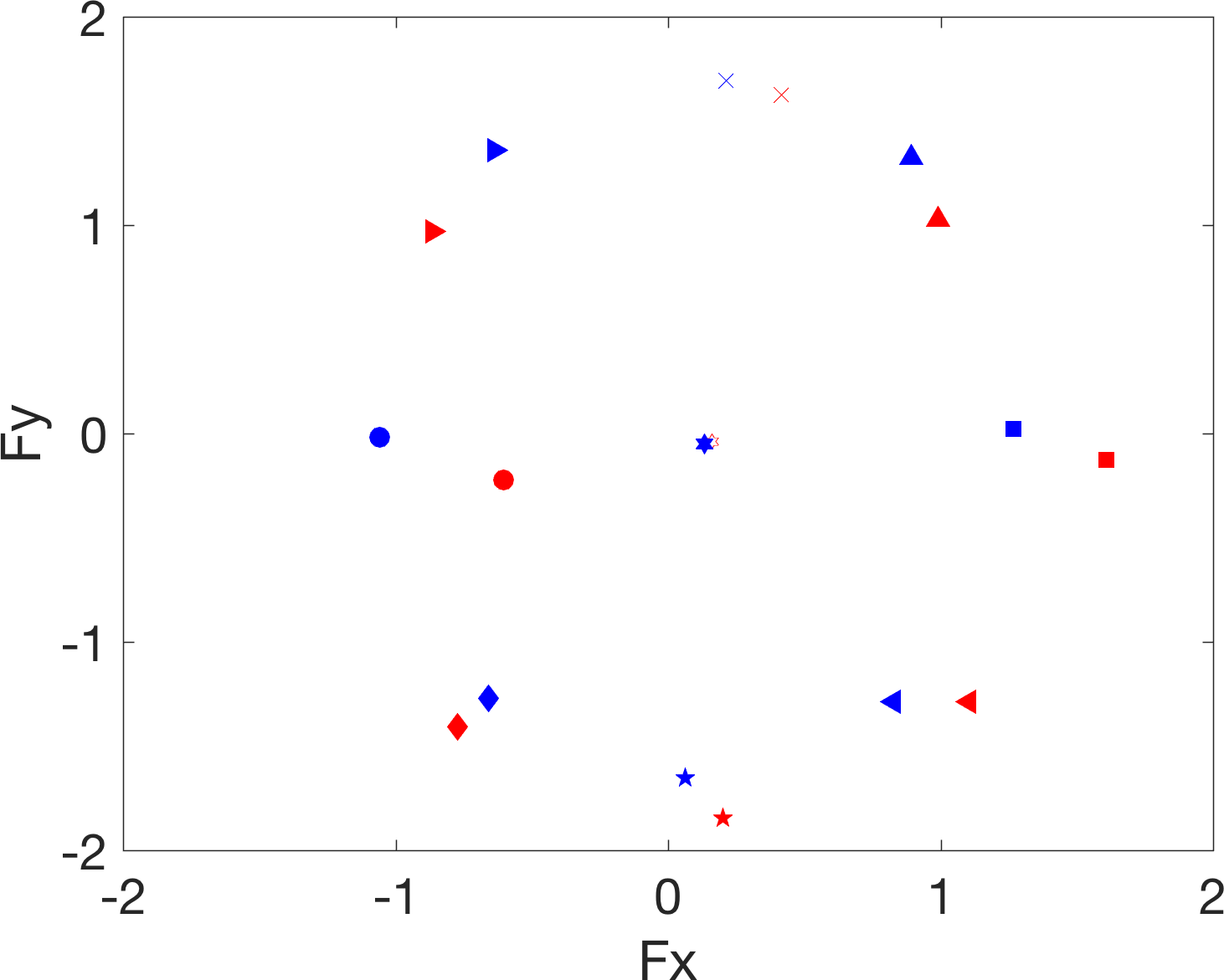}
    \end{minipage}%
  \caption{ The comparison of reconstructed resultant contact force with ground truth measured by FT sensor. (a) is for $F_x$, force in x direction; (b) is for $F_y$ and (c) is for ${F_z}$, force in normal direction. In these figures, the blue dashed line is the identity line. Figure (d) shows comparison of the force vectors in tangential plane, where blue dots represent ground truth and red dots represent the reconstructed results. } 
  \label{fig:validation}
\end{figure}

It is difficult to obtain ground truth of force distribution since most accurate sensors measure point forces instead. However, we can still perform partial validation by comparing the results of summed resultant force with fine measurement from FT sensor, which serves as a ground truth. \figref{fig:validation} shows quantitative comparison between the reconstructed force and the ground truth. The measurements of GelSlim 2.0 sensor are very close to the identity line when compared with readings from FT sensor. The standard deviation of reconstructed resultant force is $(0.244N, 0.201N, 0.322N)$ in $(x,y,z)$ directions, which is roughly within $15\%$ of the ground-truth measurement. Remaining errors are believed to be due to calibration errors, tracking errors and material non-linearity. 
The fact that correct force resultant is achieved without overfitting to experiment data shows the strength of the method, which only depends on the geometry and material parameters measured independently. Results show that the sensor can retrieve dense force distribution information with accuracy. It proves that iFEM can work well in reconstructing force from deformation measured by vision-based tactile sensors. 

\section{Conclusion}

GelSlim 2.0 is an improved version of GelSlim with a new sensing modality: dense force distribution estimation. New hardware enables this important functionality by making the sensor more rugged, improving illumination and parametrically adjustable.

Recent research has shown that vision-based sensors can provide high-resolution tactile imprints of contact via tracking of markers or calculation of optic flow on deformed gel. However, their methods lack the construction of force distribution with the measured deformation. 

The cornerstone of reconstructing force distribution with GelSlim and other vision-based tactile sensors is the mechanical modeling of the deformable gel. As a deformable continuum material, gel has infinite DOFs. While FEM, a numerical method based on continuum mechanics, can effectively model deformable objects in solving for displacements or stresses given a loading scenario.

Conversely, this paper proposes inverse FEM (iFEM), which is effective at reconstructing the external loading force based on deformation of flexible objects. This approach enables GelSlim 2.0 to provide physically realistic force distributions of contact with high spacial density. Experimental comparisons show that the integrated resultant force is fairly consistent with state of the art force-measurement.

The synthesis of hardware design and numerical methods enables the new sensor to expand the sensing capabilities of vision-based tactile sensors. The ability to effectively reconstruct force distribution could augment these experimental approaches and greatly benefit work in this area. For future work, we are interested in not only further optimization of hardware designs and numerical methods to improve accuracy, but also in exploring controls of manipulation tasks that exploit the availability of real-time dense force distribution estimation.


\bibliographystyle{bibliographies/IEEEtranN} 
{\footnotesize \bibliography{2019-ICRA-GelSlim-Force}} 

\end{document}